\documentclass{article}
\usepackage{ijcai21}

\usepackage{times}
\usepackage{soul}
\usepackage{url}
\usepackage[hidelinks]{hyperref}
\usepackage[utf8]{inputenc}
\usepackage[small]{caption}
\usepackage{graphicx}
\usepackage{amsmath}
\usepackage{amsthm}
\usepackage{booktabs}
\usepackage{algorithm}
\usepackage{algorithmic}
\usepackage{amsfonts}

\newcommand\ourmodel{MCSSL}
\newcommand{\T}{\mathcal{T}}

\newcommand{\bs}[1]{\boldsymbol{#1}}

\title{Hypernetworks for Continual Semi-Supervised Learning}

\author{
Dhanajit Brahma$^1$
\and
Vinay Kumar	Verma$^2$, \And
Piyush	Rai$^1$
\affiliations
$^1$Department of Computer Science and Engineering, IIT Kanpur, India\\
$^2$Duke University, United States
\emails
\{dhanajit, vkverma, piyush\}@cse.iitk.ac.in
}

\begin{document}

\maketitle

\begin{abstract}
  Learning from data sequentially arriving, possibly in a non i.i.d. way, with changing task distribution over time is called continual learning. 
  Much of the work thus far in continual learning focuses on supervised learning and some recent works on unsupervised learning. In many domains, each task contains a mix of labelled (typically very few) and unlabelled (typically plenty) training examples, which necessitates a semi-supervised learning approach. 
  To address this in a continual learning setting, we propose a framework for semi-supervised continual learning called Meta-Consolidation for Continual Semi-Supervised Learning (\ourmodel{}). 
  Our framework has a hypernetwork that learns the meta-distribution that generates the weights of a semi-supervised auxiliary classifier generative adversarial network (\textit{Semi-ACGAN}) as the base network. 
  We consolidate the knowledge of sequential tasks in the hypernetwork, and the base network learns the semi-supervised learning task.
  Further, we present \textit{Semi-Split CIFAR-10}, a new benchmark for continual semi-supervised learning, obtained by modifying the \textit{Split CIFAR-10} dataset, in which the tasks with labelled and unlabelled data arrive sequentially.
  Our proposed model yields significant improvements in the continual semi-supervised learning setting.
  We compare the performance of several existing continual learning approaches on the proposed continual semi-supervised learning benchmark of the Semi-Split CIFAR-10 dataset.
\end{abstract}

\section{Introduction}

Humans possess the remarkable capability of learning continuously, even in a sequential set-up. 
In machine learning, learning from data continuously arriving possibly in a non \textit{i.i.d.} way such that tasks may change over time is called continual learning, lifelong learning, or incremental learning.
Another prominent aspect of human learning is that humans do not always require supervision for the concept of an object, and they can learn by grouping similar things. In contrast, neural networks show a tendency of forgetting previously acquired knowledge when learning new tasks in a sequential manner \cite{kirkpatrick2017overcoming} which is commonly referred to as catastrophic forgetting.

With the ever-increasing diversity of data, the lack of labelled data is a common problem faced by supervised machine learning models. However, unlabelled data is plentiful and readily available to be utilized for training machine learning models. In a standard (non-continual) setting, several unsupervised learning approaches exists that can learn based on some notion of similarity without supervision. However, semi-supervised learning models can leverage both labelled and unlabeled data, thus, achieving the best of both worlds.

Most of the existing approaches for continual learning have focused on the supervised classification setting. Some recent works have explored continual unsupervised learning setting \cite{lee2019neural,rao2019continual} focusing on generative models for the image generation task.

However, most of these approaches have not investigated the semi-supervised continual learning setting. One recent work by \cite{smith2021memory} explores continual semi-supervised setting, but their setting uses the super-class structure of the CIFAR dataset, and, thus, the sequentially arriving tasks are different from our setting. Moreover, their approach uses a discriminative classifier, whereas our approach uses a generative model as the model learns the distribution of the inputs. 

Hence, we investigate a novel setting for continual semi-supervised learning where the continual learner comes across sequentially arriving tasks with labelled and unlabelled data. Similar to the standard semi-supervised learning setting, the unlabeled data and labelled data are intrinsically correlated in each learning task enabling the learner to leverage labelled and unlabelled data.

Majority of the continual learning approaches combat catastrophic forgetting by consolidating knowledge either in the weight (or parameter) space \cite{kirkpatrick2017overcoming,li2018learning,zenke2017continual,VCL}
 or in the data space \cite{AGEM,rebuffi2017icarl,shin2017continual,lopez2017gradient,aljundi2019gradient,rolnick2019experience}.
 As per the studies of the human brain, the semantic knowledge or ability to solve tasks is represented in a meta-space of high-level semantic concepts \cite{handjaras2016concepts,caramazza2003organization,mahon2009category}. Further, the memory is consolidated periodically, helping humans to learn continually \cite{caramazza1998domain,wilson1994reactivation,alvarez1994memory}. Inspired from this, recent work by \cite{joseph2020meta} proposed a framework, namely, \underline{Me}ta-Consolidation fo\underline{r} Continual \underline{L}earn\underline{in}g (MERLIN), that consolidates the knowledge of continual tasks in the meta-space, i.e., the space of the parameters of a weight generating network. This weight generating network is called the \textit{hypernetwork}, and it generates the parameters of a \textit{base network}. Such a base network is responsible for solving the specific continually arriving downstream task. We model the hypernetwork in \cite{joseph2020meta} using a Variational Auto-Encoder (VAE) model with a task-specific prior. However, they focus only on the supervised learning set-up. Thus, the base network is a discriminative neural network such as a feed-forward neural network or a modified Residual Network (ResNet-18).

In this paper, we propose \ourmodel{}: Meta-Consolidation for Continual Semi-Supervised Learning, a framework motivated from MERLIN \cite{joseph2020meta}, in which the continual learning takes place in the latent space of a weight-generating process, i.e., in the space of the parameters of the hypernetwork. However, \cite{joseph2020meta} uses a discriminative classifier (ResNets) as the base network and, thus, they focus only on the continual supervised setting. In contrast to \cite{joseph2020meta}, our model uses a modified form of an auxiliary classifier generative adversarial network (ACGAN) \cite{odena2017conditional} as the base network to perform continual semi-supervised learning. The auxiliary classifier in the GAN provides the ability to learn classification using the labelled data. Inspired from \cite{salimans2016improved}, we modify the discriminator in the ACGAN to handle the unlabelled data, and we call it \textit{Semi-ACGAN}. This leads to having a supervised and an unsupervised component in the Semi-ACGAN training objective. Thus, the VAE-like hypernetwork learns to generate parameters of the Semi-ACGAN base network, which performs the downstream task of semi-supervised classification.

\section{\ourmodel{}: Meta-Consolidation for Continual Semi-Supervised Learning}
This section starts with the problem set-up of Continual Semi-Supervised Learning. Following this, we present the overview of our proposed framework. Then, we describe the Semi-ACGAN as the base model and the training mechanism of Semi-ACGAN. Moreover, we provide the details of the hypernetwork VAE that learns the task-specific parameter distribution. Further, we describe the details of the meta-consolidation of the hypernetwork followed by the inference mechanism.

\subsection{Problem Set-up and Notation}
The problem of continual semi-supervised learning deals with learning from sequentially arriving semi-supervised tasks, as the data for a task arrives only after the previous task finishes. Let $\T_1, \T_2, \cdots, \T_{K}$ be a sequence of semi-supervised tasks such that $\T_k$ is the task at time instance $k$.
Moreover, each task $\T_j$, for $j\in\{1,\cdots,K\}$, consists of $\T_j^{tr}$, $\T_j^{val}$ and $\T_j^{test}$ that corresponds to training, validation and test sets for task $j$ respectively. Further, we define $$\T_j^{tr} = [\{(\bs{x}_m,y_m)\}_{m=1}^{M_{tr}^j}, \{\bs{u}_n\}_{n=1}^{N_{tr}^j}],$$

where $(\bs{x}_m,y_m)$ is the $m^{th}$ labelled sample, $\bs{u}_n$ is the $n^{th}$ unlabelled sample, and the total number of labelled and unlabelled training samples for $j^{th}$ task are $M_{tr}^j$ and $N_{tr}^j$ respectively.

Similarly, corresponding to the validation and test set per task, we define $\T_j^{val} = [\{(\bs{x}_m,y_m)\}_{m=1}^{M_{val}^j}, \{\bs{u}_n\}_{n=1}^{N_{val}^j}]$ and $\T_j^{test} = [\{(\bs{x}_m,y_m)\}_{m=1}^{M_{test}^j}, \{\bs{u}_n\}_{n=1}^{N_{test}^j}]$ respectively.

\subsection{Model Overview}
In our proposed framework, the hypernetwork is a VAE-like model with task-specific conditional priors, and it models the parameter distribution of the base network.
For each task, multiple instances of the base network learn the downstream semi-supervised task using both the labelled and unlabelled training data.
We use the weights of these trained base models as the inputs for training the hypernetwork.
So, the hypernetwork learns to generate task-specific weights for the base network, which eventually performs the continual semi-supervised task.
Further, meta-consolidation enables the hypernetwork to consolidate the knowledge from the previous tasks. Moreover, after training, the weights for the base network are sampled and ensembled during prediction or inference.

\begin{figure*}[!htbp]
    \centering
    \includegraphics[width=0.95\textwidth]{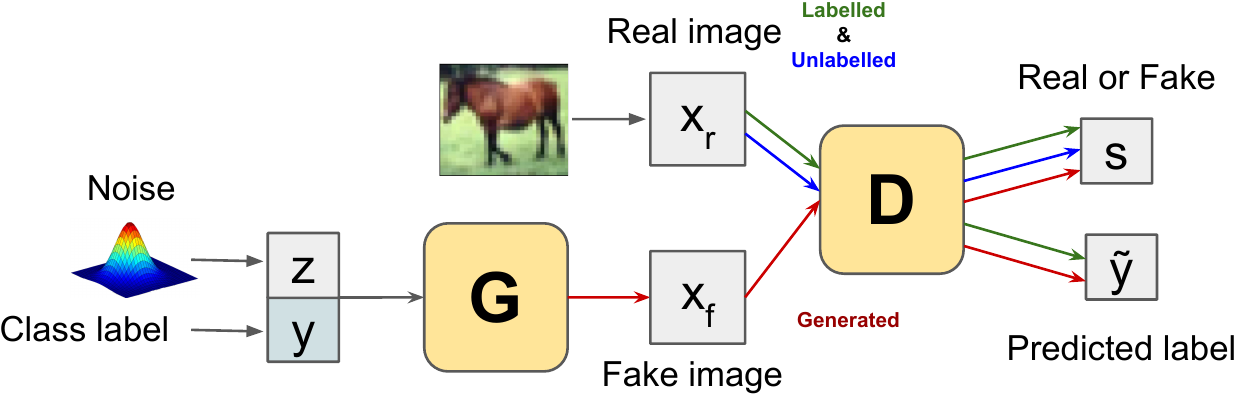}
    \caption{Base model: Semi-ACGAN. D learns from the labelled (green arrow), the unlabelled (blue arrow) samples as well as the generated (red arrow) samples. D predicts both the source (real or fake) and the class label for the labelled and generated samples, whereas, for the unlabelled samples, D only predicts the source.}\label{fig:base}
\end{figure*}

\subsection{Base Model: Semi-ACGAN}
The base network is a modified auxiliary classifier GAN (Semi-ACGAN) and, thus, consists of a generator G, a discriminator D with an auxiliary classifier.
We denote the weights of the base network for task $k$ using $\mathbf{\Theta}_k$. 

In Semi-ACGAN, G is conditioned on the class label $\bs{y}$ along with the noise $\bs{z}_b$. Thus, the generated samples $x_{\text{fake}}=\text{G}(\bs{z}_b, \bs{y})$ correspond to a class label. Let $\bs{s}$ denote whether the source of the sample $\bs{x}$ is real or fake. For a sample $\bs{x}$, the discriminator gives the probability distribution over the sources $p(\bs{s}|\bs{x})$ as well as the probability distribution over the classes $p(\bs{y}|\bs{x})$, i.e., $[p(\bs{s}|\bs{x}), p(\bs{y}|\bs{x})] = \text{D}(\bs{x})$.

Let us denote the real sample using $\bs{x}_\text{real}$ and the actual class of the sample using $\hat{\bs{y}}$. The training objective consists of the following:

(i) For Labelled data:

\hspace{0.4 cm} a. Log likelihood of the correct source,
\begin{align}
        \mathcal{L}_s^\text{L} = \mathop{\mathbb{E}}[\log p(\bs{s}=&\text{real}|\bs{x}_\text{real})] + \nonumber \\
        &\mathop{\mathbb{E}}[\log p(\bs{s}=\text{fake}|\bs{x}_\text{fake})]
\end{align}

\hspace{0.4 cm} b. Log likelihood of the correct class,
\begin{align}
        \mathcal{L}_c^\text{L} = \mathop{\mathbb{E}}[\log p(\bs{y}=&\hat{\bs{y}}|\bs{x}_\text{real})] + \nonumber \\
        &\mathop{\mathbb{E}}[\log p(\bs{y}=\hat{\bs{y}}|\bs{x}_\text{fake})]
\end{align}

(ii) For Unlabelled data:

\hspace{0.4 cm} a. Log likelihood of the correct source for real images,
\begin{align}
        \mathcal{L}_s^\text{U} = \mathop{\mathbb{E}}[\log p(\bs{s}=&\text{real}|\bs{x}_\text{real})] 
\end{align}

The discriminator D learns by maximizing $\mathcal{L}_c^\text{L}+\mathcal{L}_s^\text{L}+\mathcal{L}_s^\text{U}$, whereas the generator G learns by maximizing $\mathcal{L}_c^\text{L}-\mathcal{L}_s^\text{L}$. 

Note that since the class information of unlabelled data is missing, we do not consider the log-likelihood of the correct class in the case of unlabelled data.

The generator G is a neural network that takes both the class label and noise. The class label embedding is obtained from the class id using a class embedding layer that is trainable. Thus, G learns to generate class-specific samples.

The discriminator D is a neural network with shared layers and two separate output layers: i. validity layer: output layer for correct source, ii. auxiliary classifier layer: output layer for correct class label.
D utilizes only the validity layer for unlabelled data while using both the validity and auxiliary classifier layers for labelled data. Since the class information is known for the generated images, both the output layers of D are used for generated images.

The shared layers of D enable learning from both the labelled and the unlabelled data. As the training progresses, G learns to generate realistic samples with known class labels, enabling D to do better classification.

Figure~\ref{fig:base} shows the modules of the base network Semi-ACGAN. As the real samples can consist of both the labelled data and unlabelled data, the figure shows it using green and blue arrows, respectively. On the other hand, the red arrows depict the generated samples. Moreover, the outputs are colour coded similarly.

\subsection{Task-specific Parameter Distribution: Hypernetwork}
As $B$ instances of the trained base network are used as the inputs for training the hypernetwork, we denote this set using $\{\mathbf{\Theta}_k^l\}_{l=1}^B$ for task $k$.
Since a VAE-like model having task-specific conditional prior is used as the hypernetwork, we define the parameters of the hypernetwork as $[\bs{\theta},\bs{\phi}]$ such that $\bs{\theta}$ and $\bs{\phi}$ are the encoder and decoder parameters of the hypernetwork respectively.

The hypernetwork VAE models the task-specific parameter distribution $p(\mathbf{\Theta}|\bs{t})$. Thus, learning the hypernetwork enables the consolidation of knowledge from the previous tasks in the meta-space. 
The vector representation $\bs{t}_j$ for the $k^{th}$ task can be any fixed-length vector representation including Word2Vec \cite{mikolovdistributed}, GloVe \cite{pennington2014glove} or just a one-hot encoding of the task identity. We use $\bs{t}$ to denote $\bs{t}_j$ in this subsection for brevity. 

Inspired from MERLIN \cite{joseph2020meta}, the hypernetwork is trained by optimizing a VAE-like objective \cite{kingma2013auto}.
The computation of the marginal likelihood of the parameter distribution $p_{\bs{\theta}}(\mathbf{\Theta}|\bs{t})=\int p_{\bs{\theta}}(\mathbf{\Theta}|\bs{z},\bs{t}) p_{\bs{\theta}}(\bs{z}|\bs{t}) d\bs{z}$ is intractable because of the intractability in the computation of its true posterior $p_{\bs{\theta}}(\bs{z}|\mathbf{\Theta},\bs{t}) = \frac{p_{\bs{\theta}}(\mathbf{\Theta}|\bs{z},\bs{t}) p_{\bs{\theta}}(\bs{z}|\bs{t})}{p_{\bs{\theta}}(\mathbf{\Theta}|\bs{t})}$. Thus, we introduce an approximate variational posterior $q_{\bs{\phi}}(\bs{z}|\mathbf{\Theta},\bs{t})$ to resolve the problem of intractibility. The log marginal likelihood can be written as:
\begin{align}
    p_{\bs{\theta}}(\mathbf{\Theta}|\bs{t})= KL(q_{\bs{\phi}}(\bs{z}|\mathbf{\Theta},\bs{t})~\|~ p_{\bs{\theta}}&(\bs{z}|\mathbf{\Theta},\bs{t}))~~+ \nonumber\\ &\mathcal{L}(\bs{\theta},\bs{\phi}|\mathbf{\Theta},\bs{t})
\end{align}

where $\mathcal{L}(\bs{\theta},\bs{\phi}|\mathbf{\Theta},\bs{t}) = \int_{\bs{z}} q_{\bs{\phi}}(\bs{z}|\mathbf{\Theta},\bs{t}) \log\frac{p_{\bs{\theta}}(\bs{z},\mathbf{\Theta}|\bs{t})}{q_{\bs{\phi}}(\bs{z}|\mathbf{\Theta},\bs{t})} $ is the evidence lower bound (ELBO). This lower bound can be maximized in order to maximize the log-likelihood.

Further, $\mathcal{L}(\bs{\theta},\bs{\phi}|\mathbf{\Theta},\bs{t})$ can be expressed as (refer to \cite{joseph2020meta} for complete derivation):
\begin{align}
    \mathcal{L}(\bs{\theta},\bs{\phi}|\mathbf{\Theta},\bs{t}) = -KL(q_{\bs{\phi}}&(\bs{z}|\mathbf{\Theta},\bs{t})~\|~ p_{\bs{\theta}}(\bs{z}|\bs{t}))~~+ \nonumber \\
    & \mathop{\mathbb{E}}_{q_{\bs{\phi}}(\bs{z}|\mathbf{\Theta},\bs{t})}[\log p_{\bs{\theta}}(\mathbf{\Theta}|\bs{z},\bs{t})]
    \label{eq5}
\end{align}
Maximizing Eqn.~\ref{eq5} minimizes the KL divergence term, causing the approximate posterior weights to become close to the task-specific prior $p_{\bs{\theta}}(\bs{z}|\bs{t})$. The second term is the expected negative reconstruction error, and it requires sampling to estimate. The hypernetwork parameters $\bs{\phi}$ and $\bs{\theta}$, also known as encoder and decoder parameters, are trained using backpropagation and stochastic gradient descent. We assume that $p_{\bs{\theta}}(.)$ and $q_{\bs{\phi}}(.)$ are Gaussian distributions. Moreover, the reparameterization trick \cite{kingma2013auto} is used to backpropagate through the stochastic parameters. Taking $\{\mathbf{\Theta}_k^l\}_{l=1}^B$ as the input, we train the hypernetwork by maximizing Eqn.~\ref{eq5}. 

Unlike standard VAE, the task-specific prior is not an isotropic multivariate Gaussian. It is given by:
\begin{align}
    p_{\bs{\theta}}(\bs{z}|\bs{t})
    = \mathcal{N}(\bs{z}|\bs{\mu_{\bs{t}}},\bs{\Sigma_{\bs{t}}})
\end{align}

where $\bs{\mu_{\bs{t}}}=\bs{W}_{\bs{\mu}}^T\bs{t}$ and
$\bs{\Sigma_{\bs{t}}}=\bs{W}_{\bs{\Sigma}}^T \bs{t}$ such that
$\bs{W}_{\bs{\mu}}$ and $\bs{W}_{\bs{\Sigma}}$ are trainable parameters, and learned along with the hypernetwork parameters.

\subsection{Meta-Consolidation} 
Training the VAE directly on $\{\mathbf{\Theta}_k^l\}_{l=1}^B$ causes a distributional shift, i.e., a bias towards the current task $k$.
So, the hypernetwork VAE needs to consolidate the knowledge from the previous tasks. We call this \textit{meta-consolidation}.
We store the means and covariances of all the learned task-specific prior, which adds a negligible storage complexity.
The meta-consolidation mechanism is described below:
\begin{enumerate}
    \item For each task $\T_j$ till current task $k$ ($j={1,\cdots,k}$),
    \begin{enumerate}
        \item Sample $\bs{z_{t_j}}$ from task-specific prior:
        $$\bs{z_{t_j}} \sim \mathcal{N}(\bs{z}|\bs{\mu_{\bs{t_j}}},\bs{\Sigma_{\bs{t_j}}})$$
        \item Sample $P$ many semi-supervised base pseudo-models from decoder:
        $$\mathbf{\Theta}_j^i \sim p_{\bs{\theta}}(\mathbf{\Theta}|\bs{z_{t_j}},\bs{t}_j);~ \text{where}~i\in\{1,2,\cdots,P\}$$
        \item Compute the loss using Eqn.~\ref{eq5}:
            $$Loss = \sum_{i=1}^P \mathcal{L}(\bs{\theta},\bs{\phi}|\mathbf{\Theta}_j^i,\bs{t}_j)$$
        \item Optimize $Loss$ to update parameters $\bs{\phi}$, $\bs{\theta}$
    \end{enumerate}
\end{enumerate} 

\subsection{Inference}
Learning the task-specific parameter distribution $p_{\bs{\theta}}(\mathbf{\Theta}|\bs{z},\bs{t})$ gives the ability to sample multiple $\mathbf{\Theta}$'s during inference. This ability provides an \textit{ensembling} effect of multiple models without storing the models a priori. Like most of the other continual learning approaches, we use a small exemplar memory buffer $\mathcal{E}$ for fine-tuning during inference. 

Our approach can work with or without task-specific information during inference. However, we focus on the task-agnostic setting as it is more realistic and challenging.
The inference procedure for task-agnostic inference is described below:
\begin{enumerate}
    \item Aggregate the stored means and covariances:
$$ \bs{\mu} = \frac{1}{k}\sum_{i=1}^k \bs{\mu}_{t_i},
\bs{\Sigma} = \frac{1}{k}\sum_{i=1}^k \bs{\Sigma}_{t_i}$$
    \item Sample $\bs{z}$ from prior with aggregated mean and covariance:
$$\bs{z} \sim \mathcal{N}(\bs{z}|\bs{\mu},\bs{\Sigma})$$
    \item Sample $E$ number of $\mathbf{\Theta}$'s (semi-supervised base models) from learned decoder: 
$$ \mathbf{\Theta}^i \sim p_{\bs{\theta}}(\mathbf{\Theta}|\bs{z});~ \text{where}~i\in\{1,2,\cdots,E\}$$
    \item Fine-tune $\{\mathbf{\Theta}^i\}_{i=1}^E$ on $\mathcal{E}$
    \item Ensemble results from $\{\mathbf{\Theta}^i\}_{i=1}^E$ and solve tasks $\{{\T_i}\}_{i=1}^k$
\end{enumerate}
\vspace{0.5cm}
The inference procedure for task-aware inference is given as below:
\begin{enumerate}
    \item For each task $\T_j;$ $j\in\{1,\cdots,k\}$
    \begin{enumerate}
        \item Sample $\bs{z_{t_j}}$ from task-specific prior:
    $$\bs{z_{t_j}} \sim \mathcal{N}(\bs{z}|\bs{\mu_{t_j}},\bs{\Sigma_{t_j}})$$
        \item Sample $E$ number of $\mathbf{\Theta}$'s from learned decoder: 
    $$ \mathbf{\Theta}_j^{i} \sim p_{\bs{\theta}}(\mathbf{\Theta}|\bs{z_{t_j}},\bs{t_j});~ \text{where}~i\in\{1,2,\cdots,E\}$$
        \item Fine-tune $\{\mathbf{\Theta}_j^{i}\}_{i=1}^E$ on $\mathcal{E}$
        \item Ensemble results from $\{\mathbf{\Theta}_j^{i}\}_{i=1}^E$ and solve task $\T_j$
        \end{enumerate}
\end{enumerate}

\begin{table*}[!htbp]
\small
    \centering
    \resizebox{0.90\textwidth}{!}{
    \addtolength{\tabcolsep}{4pt}
    \begin{tabular}{|c|cc|cc|cc|cc|cc|cc|} \toprule
        \multicolumn{1}{c}{\textbf{Labelled data$\rightarrow$}} & \multicolumn{2}{c}{\textbf{500 (120)}} & \multicolumn{2}{c}{\textbf{250 (60)}} & \multicolumn{2}{c}{\textbf{100 (24)}} & \multicolumn{2}{c}{\textbf{50 (12)}} \\ 
        \cmidrule(l{2pt}r{2pt}){2-3} \cmidrule(l{2pt}r{2pt}){4-5} \cmidrule(l{2pt}r{2pt}){6-7} \cmidrule(l{2pt}r{2pt}){8-9}
        {\textbf{Methods$\downarrow$}} & $\bs{A\uparrow}$ & $\bs{F\downarrow}$ & $\bs{A\uparrow}$ & $\bs{F\downarrow}$ &  $\bs{A\uparrow}$ & $\bs{F\downarrow}$& $\bs{A\uparrow}$ & $\bs{F\downarrow}$\\ 
        \hline
        {Single-SSL}  & {59.57}& {2.90}&  {59.80}& {3.44}&  {57.35}& \textbf{1.87}&  {54.38}& \textbf{0.72}\\
        {EWC-SSL}  & {60.11}& {0.88}&  {60.89}& {3.20}&  {60.78}& {3.17}&  {59.65}& {2.23}\\
		\ourmodel{} (Ours) & \textbf{63.72}& \textbf{-4.61}&  \textbf{63.55}& \textbf{2.45}& \textbf{63.37}& {3.60}& \textbf{63.00}& {11.25}\\
        \bottomrule
    \end{tabular}
    }
    \caption{Average accuracy ($A$) ($\uparrow$ higher is better) and average forgetting ($F$) ($\downarrow$ lower is better) of \ourmodel{} and the baseline approaches modified to use both labelled and unlabelled data on Semi-Split CIFAR-10 dataset. The number of labelled data is varied, whereas, the number of unlabelled data is fixed to 1000 (240). The number inside () denotes the labelled memory buffer size.}
    \label{base-ssl-results}

\end{table*}

\begin{table*}[!htbp]
\small
    \centering
    \resizebox{0.90\textwidth}{!}{
    \addtolength{\tabcolsep}{4pt}
    \begin{tabular}{|c|cc|cc|cc|cc|cc|cc|} \toprule
        \multicolumn{1}{c}{\textbf{Labelled data$\rightarrow$}} & \multicolumn{2}{c}{\textbf{500 (120)}} & \multicolumn{2}{c}{\textbf{250 (60)}} & \multicolumn{2}{c}{\textbf{100 (24)}} & \multicolumn{2}{c}{\textbf{50 (12)}} \\ 
        \cmidrule(l{2pt}r{2pt}){2-3} \cmidrule(l{2pt}r{2pt}){4-5} \cmidrule(l{2pt}r{2pt}){6-7} \cmidrule(l{2pt}r{2pt}){8-9}
        {\textbf{Methods$\downarrow$}} & $\bs{A\uparrow}$ & $\bs{F\downarrow}$ & $\bs{A\uparrow}$ & $\bs{F\downarrow}$ &  $\bs{A\uparrow}$ & $\bs{F\downarrow}$& $\bs{A\uparrow}$ & $\bs{F\downarrow}$\\ 
        \hline
        {Single}  & {63.48}& {10.81}&  {59.54}& {1.81}&  {53.65}& \textbf{-3.76}&  {51.68}& {3.57}\\
        {EWC}  & \textbf{67.32}& {5.20}&  {59.25}& {10.01}&  {52.82}& {-0.39}&  {49.82}& {0.04}\\
        {MERLIN} & {59.71}& {0.51}&  {33.77}& \textbf{-0.48}&  {22.71}&  {2.07}& {17.84}& {1.06}\\
		\ourmodel{} (Ours) & {63.72}& \textbf{-4.61}&  \textbf{63.55}& {2.45}& \textbf{63.37}& {3.60}& \textbf{63.00}& {11.25}\\
        \bottomrule
    \end{tabular}
    }
    \caption{Average accuracy ($A$) ($\uparrow$ higher is better) and average forgetting ($F$) ($\downarrow$ lower is better) upon varying the number of labelled data on Semi-Split CIFAR-10 dataset. The number inside () denotes the labelled memory buffer size. \ourmodel{} (Ours) uses some unlabelled data, whereas other baseline models only use the labelled data.}
    \label{csl-mcssl}

\end{table*}

\section{Related Work}
Most of the existing continual learning approaches focus on the problem of continual supervised learning. These approaches consolidate knowledge either in the weight space, data space or meta-space. 

Elastic Weight Consolidation (EWC) \cite{kirkpatrick2017overcoming} is a regularization based approach that penalizes drastic changes in the parameters that have a large influence on prediction. Variational Continual Learning (VCL) \cite{VCL} is a probabilistic regularization based approach using Bayesian neural networks. They treat the posterior of the current task as the prior for the next task as it naturally emerges from online variational inference. 
Learning without Forgetting (LwF) \cite{li2018learning} uses knowledge distillation to preserve the knowledge of previous tasks.

Gradient Episodic Memory (GEM) \cite{lopez2017gradient} stores a limited number of samples to retrain while constraining new task updates to not interfere with knowledge of previous tasks.  \cite{aljundi2019gradient} extended this idea by selecting subsets of samples that approximate the region of the data seen in the previous tasks.

\cite{von2019continual} operates in the meta-space as it learns a hypernetwork that generates the weights of the base model. However, they learn a task identity conditioned deterministic function.
Similarly, recent work by \cite{joseph2020meta} consolidates the knowledge from previous tasks in the meta-space of weight generating hypernetwork. They learn the task-specific distribution of weights, giving them the ability to ensemble during prediction.

Some recent approaches focus on the problem of continual unsupervised learning. 
\cite{rao2019continual} presents an approach for unsupervised representation learning with a dynamic expansion based approach using a latent mixture-of-Gaussians. 
\cite{lee2019neural} focuses on the discriminative and generative tasks using Dirichlet process mixture models for dynamic expansion with a generative process different from \cite{rao2019continual}.

A recent approach, named, \textit{DistillMatch} \cite{smith2021memory} tries to address the problem of Continual Semi-Supervised Learning. However, the unlabeled data in continual tasks significantly differ from our set-up, as \cite{smith2021memory} uses the super-class structure of the CIFAR dataset. 
DistillMatch combines pseudo-labelling for semi-supervised learning,
knowledge distillation for continual learning,
along with consistency regularization as it uses \textit{FixMatch} \cite{sohn2020fixmatch} as the base semi-supervised learner.
Moreover, it has an out-of-distribution detection scheme required due to its problem set-up.
However, unlike our approach, DistillMatch is not a generative approach, and thus, the distribution of input is not directly modeled.

\section{Experiments}
We propose a novel dataset for continual semi-supervised learning. We conduct a comprehensive analysis of our proposed model \ourmodel{} on the proposed modified CIFAR dataset and compare our model with other state-of-the-art approaches. We describe the dataset details, evaluation metrics, hyperparameter settings, experimental results and analysis in this section. All the results are shown for \textit{task-agnostic} setting as it is more realistic and challenging.

\begin{figure}[!hb]
	\centering
	\includegraphics[scale=0.48]{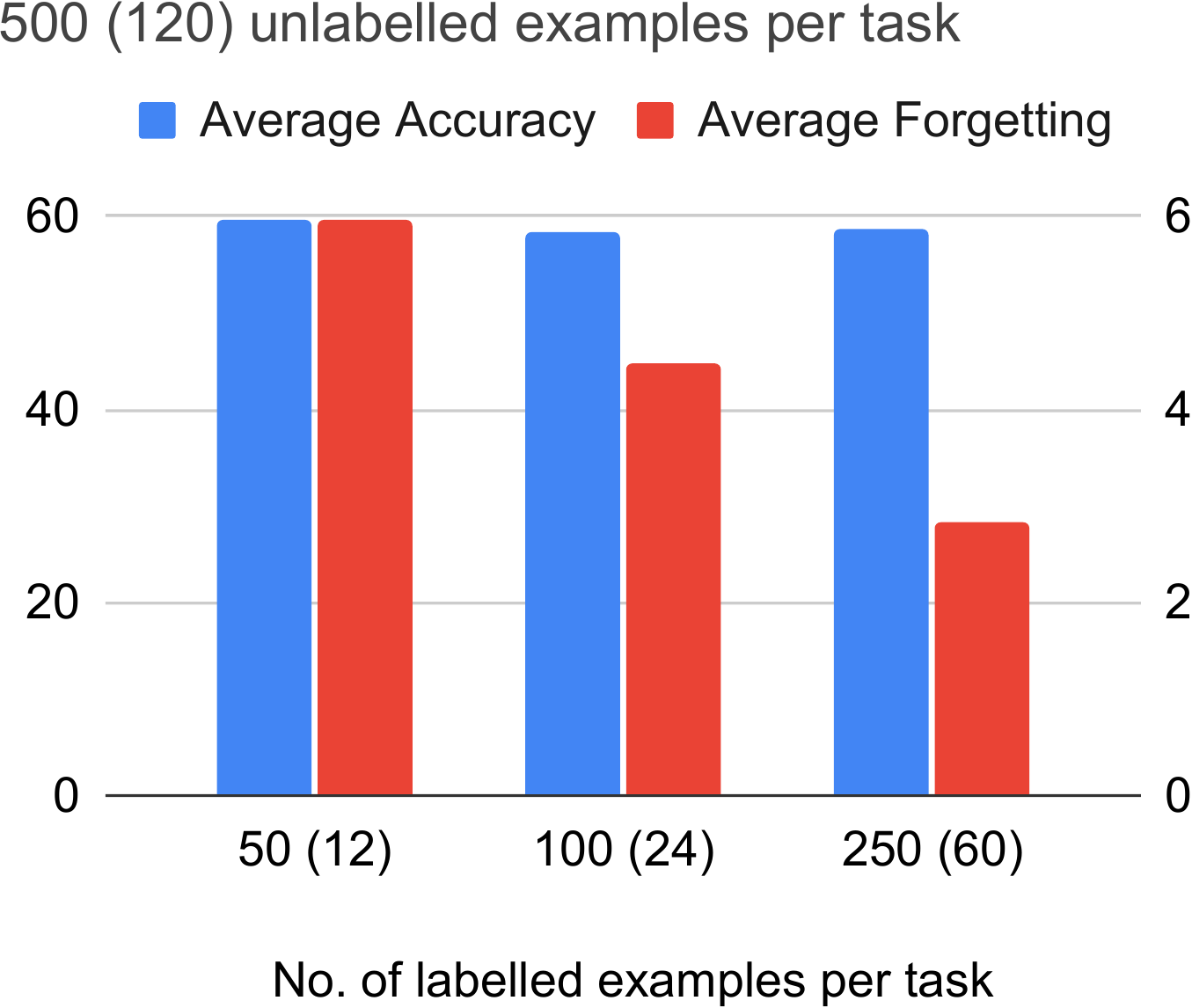}

	\vspace{0.2cm}	
	
	\includegraphics[scale=0.48]{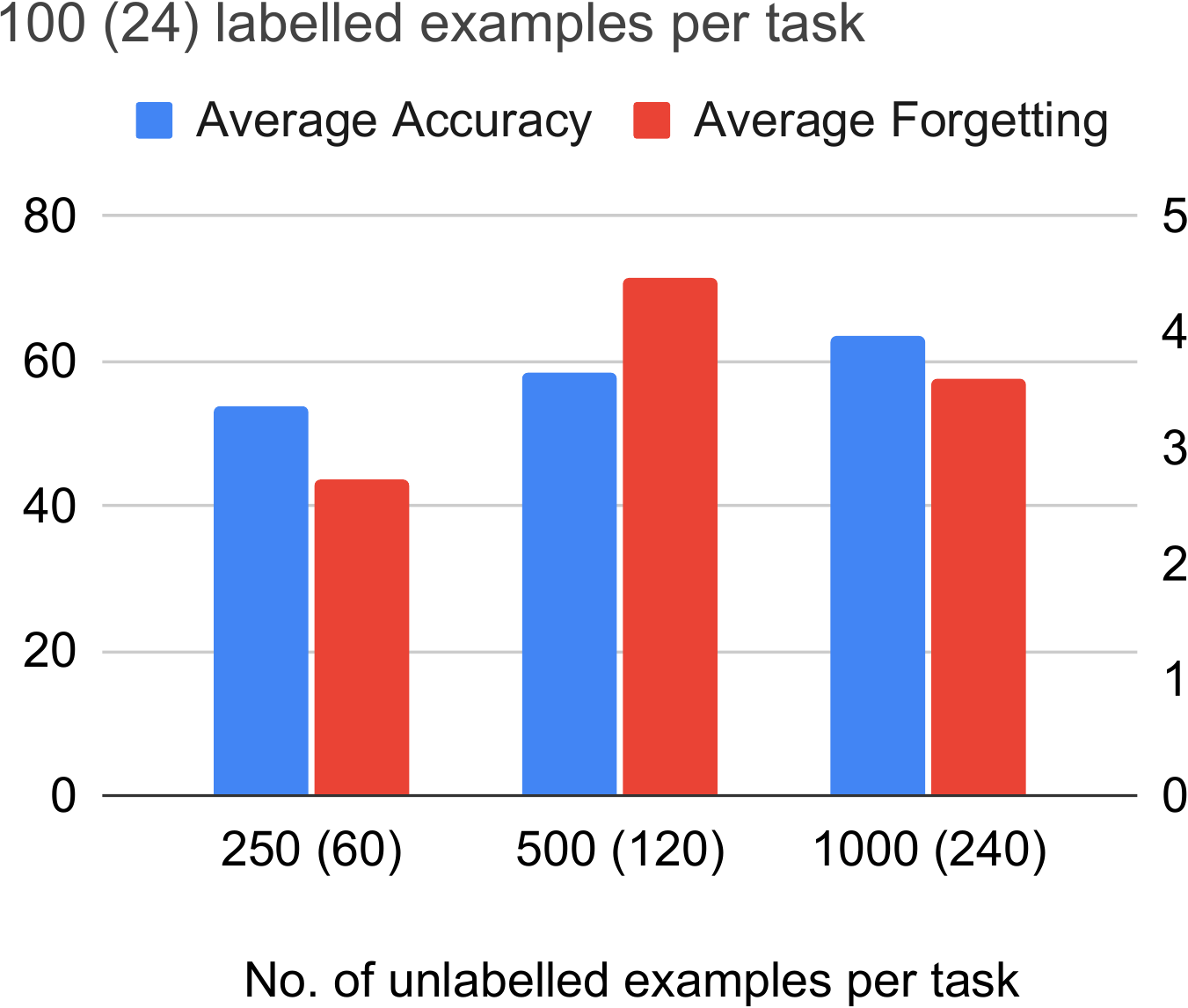}
	\caption{\textbf{Top:} Semi-Split CIFAR-10 dataset with varying number of labelled data, and fixed number of unlabelled data. \textbf{Bottom:} Semi-Split CIFAR-10 dataset with varying number of unlabelled data, and fixed number of labelled data. The number inside () denotes memory buffer size.}
	\label{fig:varying}
\end{figure}

\subsection{Dataset Details}
We experiment with the CIFAR dataset in a continual learning semi-supervised set-up. \textit{Split CIFAR-10 }\cite{zenke2017continual} dataset is a supervised continual learning benchmark dataset that has 10 tasks with 45,000 images in total such that each task contains 2 classes.
We modify Split CIFAR-10 dataset for continual semi-supervised learning set-up. Thus, we have 10 tasks in total where each task contains 2 classes with a varying number of labelled data and unlabelled data. We name this modified dataset as \textit{Semi-Split CIFAR-10}.

\subsection{Training Details and Hyperparameters}
The base model Semi-ACGAN uses convolutional deep neural networks for G and D. The validity and auxiliary classifier layers of D are both linear layers with sigmoid and softmax functions applied respectively to get the scores. During training, number of base models is 5, and for training these base models, we use Adam optimizer with initial learning rate of 0.0002. We provide the architecture details of G and D below.

Detailed architecture of G: [\texttt{BatchNorm}; \texttt{Upsampling}: scale=4; \texttt{Conv}: 3x3, 16 filters, stride=1, padding=1); \texttt{BatchNorm}; \texttt{LeakyReLU}; \texttt{Conv}: 3x3, 3 filters, stride=1, padding=1); \texttt{Tanh}].

Detailed architecture of shared layers of D: [\texttt{Conv}: 3x3, 16 filters, stride=2, padding=1; \texttt{LeakyReLU}; \texttt{Dropout}: p=0.25; \texttt{Conv}: 3x3, 32 filters, stride=2, padding=1; \texttt{LeakyReLU}; \texttt{Dropout}: p=0.25; \texttt{BatchNorm}].

The hypernetwork uses 5 base models to learn its encoder and decoder parameters. The chunking trick proposed by \cite{von2019continual} is used to keep the size of the VAE small. The weights of the base models are flattened and then split into chunks of size 250. We train the hypernetwork conditioned on the chunks, and the chunk embeddings are learned together with the hypernetwork parameters. The weights are assembled back for making the inference. We use the one-hot encoding of the task identity. 

In the hypernetwork VAE, the encoder network has one fully connected layer with 30 neurons. Moreover, the decoder network architecture is a mirror of the encoder network architecture. This is followed by two layers for predicting mean and diagonal covariance vectors respectively. The dimension of latent $\bs{z}$ is 10. The hypernetwork VAE is trained for 5 epochs using Adadelta optimizer with an initial learning rate of 0.005 and batch size of 1.

During inference, we sample 15 models from the decoder for ensembling using majority voting. Further, the labelled and unlabelled data in the memory buffer are used to fine-tune the models. Since the base models are loaded sequentially at a time (saving only the final logits) not more than one model is ever loaded in the memory at one time.

\subsection{Evaluation Metrics}
We define $a_{k,j}$ as the accuracy on the test set of $j^{th}$ task, after model is trained on $k^{th}$ task.
Following previous works on continual learning, we use the metrics given below to evaluate the models:
\begin{enumerate}
	\item Average Accuracy: 
	$$ A = \frac{1}{K}\sum_{k=1}^K A_k~; \text{ where }A_k = \frac{1}{k}\sum_{j=1}^k a_{k,j}$$
	\item Average Forgetting:
	\begin{align*}
		&F = \frac{1}{K}\sum_{k=1}^K F_k~; \\
	\text{ where }F_k &= \frac{1}{k-1}\sum_{j=1}^{k-1} \max_{l\in\{1,\cdots,k-1\}} (a_{l,j}-a_{k,j})
	\end{align*}	 
\end{enumerate}

\subsection{Results and Analysis}
We adapt EWC for continual semi-supervised setting, denoted as \textit{EWC-SSL}, and conduct experiments with a varying number of labelled data. We also train a single base model Semi-ACGAN without any continual learning mechanism and denote it using \textit{Single-SSL}.
The labelled data is used during continual training, and a small fraction of it is stored in a memory buffer for fine-tuning during inference. Table~\ref{base-ssl-results} shows the results on Semi-Split CIFAR-10 dataset. We compare our approach with Single-SSL and EWC-SSL as baseline continual semi-supervised approaches. We fix number of unlabelled data to 1000 per task with unlabelled data memory buffer size of 240 per task for all the models. The decrease in number of labelled data does not have much significant impact on the performance of \ourmodel{} as it consistently outperforms baseline approaches in all settings. This suggests that our approach generalises better in low data regimes.

In order to demonstrate the ability of our model to leverage unlabeled data, we also compare with other continual supervised baseline approaches in Table~\ref{csl-mcssl}. Here, \textit{Single} denotes a \textit{Resnet18} classifier trained without any continual learning mechanism. EWC also uses a Resnet18 classifier, whereas MERLIN uses a modified Resnet18 classifier as described in \cite{joseph2020meta}. Our model uses labelled data along with some unlabelled data, whereas other models use only labelled data. As our model outperforms other approaches in most settings, we observe that it does better than others as labelled data decreases. Decreasing labelled data has no significant effect on our model, whereas the performance of other models drops drastically.

In Fig.~\ref{fig:varying} (Top), we fix the number of unlabelled examples per task as 500 with 120 unlabelled samples in the memory buffer for fine-tuning as we vary the number of labelled examples per task. We notice that the accuracy slightly increases with an increase in the labelled data.
Here, forgetting tends to decrease as the labelled examples per task increases.

Fig.~\ref{fig:varying} (Bottom) shows a varying number of unlabelled examples upon fixing the number of labelled examples per task as 100 with 24 labelled examples in the memory buffer for fine-tuning. We observe that upon fixing the number of labelled data per task, the accuracy increases with an increase in unlabelled data.
Also, forgetting tends to increase as the examples per task increases, but it is not significant.

\section{Conclusion}
We proposed a novel continual semi-supervised learning scheme in which tasks with both unlabeled and labelled examples arrive sequentially. We developed a task-specific weight generation-based approach for continual semi-supervised learning problem. We utilize a semi-supervised auxiliary classifier GAN (\textit{Semi-ACGAN}) as the base model. We also extended other continual learning approaches to use both labelled and unlabelled data, and comparisons show that \ourmodel{} performs better on the \textit{Semi-Split CIFAR-10} dataset in most of the settings. We also outperform other continual supervised baseline approaches that show the ability of our model \ourmodel{} to leverage knowledge from the unlabelled examples. Moreover, \ourmodel{} performs well even for a low number of labelled examples. In future work, we plan to evaluate \ourmodel{} on more benchmarks and baselines.

\textbf{Acknowledgments:} This work was supported by Qualcomm Innovation Fellowship.

\bibliographystyle{named}
\bibliography{ijcai21}

\begin{thebibliography}{}

\bibitem[\protect\citeauthoryear{Aljundi \bgroup \em et al.\egroup
  }{2019}]{aljundi2019gradient}
Rahaf Aljundi, Min Lin, Baptiste Goujaud, and Yoshua Bengio.
\newblock Gradient based sample selection for online continual learning.
\newblock In {\em NeurIPS}, 2019.

\bibitem[\protect\citeauthoryear{Alvarez and Squire}{1994}]{alvarez1994memory}
Pablo Alvarez and Larry~R Squire.
\newblock Memory consolidation and the medial temporal lobe: a simple network
  model.
\newblock {\em PNAS}, 91(15), 1994.

\bibitem[\protect\citeauthoryear{Caramazza and
  Mahon}{2003}]{caramazza2003organization}
Alfonso Caramazza and Bradford~Z Mahon.
\newblock The organization of conceptual knowledge: the evidence from
  category-specific semantic deficits.
\newblock {\em Trends in cognitive sciences}, 7(8), 2003.

\bibitem[\protect\citeauthoryear{Caramazza and
  Shelton}{1998}]{caramazza1998domain}
Alfonso Caramazza and Jennifer~R Shelton.
\newblock Domain-specific knowledge systems in the brain: The animate-inanimate
  distinction.
\newblock {\em Journal of cognitive neuroscience}, 10(1), 1998.

\bibitem[\protect\citeauthoryear{Chaudhry \bgroup \em et al.\egroup
  }{2019}]{AGEM}
Arslan Chaudhry, Marc’Aurelio Ranzato, Marcus Rohrbach, and Mohamed
  Elhoseiny.
\newblock Efficient lifelong learning with a-gem.
\newblock In {\em ICLR}, 2019.

\bibitem[\protect\citeauthoryear{Handjaras \bgroup \em et al.\egroup
  }{2016}]{handjaras2016concepts}
Giacomo Handjaras, Emiliano Ricciardi, Andrea Leo, Alessandro Lenci, Luca
  Cecchetti, Mirco Cosottini, Giovanna Marotta, and Pietro Pietrini.
\newblock How concepts are encoded in the human brain: a modality independent,
  category-based cortical organization of semantic knowledge.
\newblock {\em Neuroimage}, 135, 2016.

\bibitem[\protect\citeauthoryear{Joseph and
  Balasubramanian}{2020}]{joseph2020meta}
KJ~Joseph and Vineeth~N Balasubramanian.
\newblock Meta-consolidation for continual learning.
\newblock {\em arXiv preprint arXiv:2010.00352}, 2020.

\bibitem[\protect\citeauthoryear{Kingma and Welling}{2013}]{kingma2013auto}
Diederik~P Kingma and Max Welling.
\newblock Auto-encoding variational bayes.
\newblock {\em arXiv preprint arXiv:1312.6114}, 2013.

\bibitem[\protect\citeauthoryear{Kirkpatrick \bgroup \em et al.\egroup
  }{2017}]{kirkpatrick2017overcoming}
James Kirkpatrick, Razvan Pascanu, Neil Rabinowitz, Joel Veness, Guillaume
  Desjardins, Andrei~A Rusu, Kieran Milan, John Quan, Tiago Ramalho, Agnieszka
  Grabska-Barwinska, et~al.
\newblock Overcoming catastrophic forgetting in neural networks.
\newblock {\em Proceedings of the national academy of sciences}, 114(13), 2017.

\bibitem[\protect\citeauthoryear{Lee \bgroup \em et al.\egroup
  }{2019}]{lee2019neural}
Soochan Lee, Junsoo Ha, Dongsu Zhang, and Gunhee Kim.
\newblock A neural dirichlet process mixture model for task-free continual
  learning.
\newblock In {\em ICLR}, 2019.

\bibitem[\protect\citeauthoryear{Li and Hoiem}{2018}]{li2018learning}
Zhizhong Li and Derek Hoiem.
\newblock Learning without forgetting.
\newblock {\em IEEE TPAMI}, 40(12), 2018.

\bibitem[\protect\citeauthoryear{Lopez-Paz and
  Ranzato}{2017}]{lopez2017gradient}
David Lopez-Paz and Marc'Aurelio Ranzato.
\newblock Gradient episodic memory for continual learning.
\newblock In {\em NeurIPS}, 2017.

\bibitem[\protect\citeauthoryear{Mahon \bgroup \em et al.\egroup
  }{2009}]{mahon2009category}
Bradford~Z Mahon, Stefano Anzellotti, Jens Schwarzbach, Massimiliano Zampini,
  and Alfonso Caramazza.
\newblock Category-specific organization in the human brain does not require
  visual experience.
\newblock {\em Neuron}, 63(3), 2009.

\bibitem[\protect\citeauthoryear{Mikolov \bgroup \em et al.\egroup
  }{2013}]{mikolovdistributed}
Tomas Mikolov, Ilya Sutskever, Kai Chen, Greg Corrado, and Jeffrey Dean.
\newblock Distributed representations of words and phrases and their
  compositionality.
\newblock {\em NeurIPS}, 2013.

\bibitem[\protect\citeauthoryear{Nguyen \bgroup \em et al.\egroup }{2018}]{VCL}
Cuong~V Nguyen, Yingzhen Li, Thang~D Bui, and Richard~E Turner.
\newblock Variational continual learning.
\newblock In {\em ICLR}, 2018.

\bibitem[\protect\citeauthoryear{Odena \bgroup \em et al.\egroup
  }{2017}]{odena2017conditional}
Augustus Odena, Christopher Olah, and Jonathon Shlens.
\newblock Conditional image synthesis with auxiliary classifier gans.
\newblock In {\em International conference on machine learning}. PMLR, 2017.

\bibitem[\protect\citeauthoryear{Pennington \bgroup \em et al.\egroup
  }{2014}]{pennington2014glove}
Jeffrey Pennington, Richard Socher, and Christopher~D Manning.
\newblock Glove: Global vectors for word representation.
\newblock In {\em EMNLP}, 2014.

\bibitem[\protect\citeauthoryear{Rao \bgroup \em et al.\egroup
  }{2019}]{rao2019continual}
Dushyant Rao, Francesco Visin, Andrei Rusu, Razvan Pascanu, Yee~Whye Teh, and
  Raia Hadsell.
\newblock Continual unsupervised representation learning.
\newblock {\em NeurIPS}, 32, 2019.

\bibitem[\protect\citeauthoryear{Rebuffi \bgroup \em et al.\egroup
  }{2017}]{rebuffi2017icarl}
Sylvestre-Alvise Rebuffi, Alexander Kolesnikov, Georg Sperl, and Christoph~H
  Lampert.
\newblock icarl: Incremental classifier and representation learning.
\newblock In {\em CVPR}. IEEE, 2017.

\bibitem[\protect\citeauthoryear{Rolnick \bgroup \em et al.\egroup
  }{2019}]{rolnick2019experience}
David Rolnick, Arun Ahuja, Jonathan Schwarz, Timothy Lillicrap, and Gregory
  Wayne.
\newblock Experience replay for continual learning.
\newblock In {\em NeurIPS}, 2019.

\bibitem[\protect\citeauthoryear{Salimans \bgroup \em et al.\egroup
  }{2016}]{salimans2016improved}
Tim Salimans, Ian~J Goodfellow, Wojciech Zaremba, Vicki Cheung, Alec Radford,
  and Xi~Chen.
\newblock Improved techniques for training gans.
\newblock In {\em NeurIPS}, 2016.

\bibitem[\protect\citeauthoryear{Shin \bgroup \em et al.\egroup
  }{2017}]{shin2017continual}
Hanul Shin, Jung~Kwon Lee, Jaehong Kim, and Jiwon Kim.
\newblock Continual learning with deep generative replay.
\newblock In {\em NeurIPS}, 2017.

\bibitem[\protect\citeauthoryear{Smith \bgroup \em et al.\egroup
  }{2021}]{smith2021memory}
James Smith, Jonathan Balloch, Yen-Chang Hsu, and Zsolt Kira.
\newblock Memory-efficient semi-supervised continual learning: The world is its
  own replay buffer.
\newblock {\em arXiv preprint arXiv:2101.09536}, 2021.

\bibitem[\protect\citeauthoryear{Sohn \bgroup \em et al.\egroup
  }{2020}]{sohn2020fixmatch}
Kihyuk Sohn, David Berthelot, Nicholas Carlini, Zizhao Zhang, Han Zhang,
  Colin~A Raffel, Ekin~Dogus Cubuk, Alexey Kurakin, and Chun-Liang Li.
\newblock Fixmatch: Simplifying semi-supervised learning with consistency and
  confidence.
\newblock {\em NeurIPS}, 33, 2020.

\bibitem[\protect\citeauthoryear{von Oswald \bgroup \em et al.\egroup
  }{2019}]{von2019continual}
Johannes von Oswald, Christian Henning, Jo{\~a}o Sacramento, and Benjamin~F
  Grewe.
\newblock Continual learning with hypernetworks.
\newblock In {\em ICLR}, 2019.

\bibitem[\protect\citeauthoryear{Wilson and
  McNaughton}{1994}]{wilson1994reactivation}
Matthew~A Wilson and Bruce~L McNaughton.
\newblock Reactivation of hippocampal ensemble memories during sleep.
\newblock {\em Science}, 265(5172), 1994.

\bibitem[\protect\citeauthoryear{Zenke \bgroup \em et al.\egroup
  }{2017}]{zenke2017continual}
Friedemann Zenke, Ben Poole, and Surya Ganguli.
\newblock Continual learning through synaptic intelligence.
\newblock In {\em ICML}. JMLR. org, 2017.

\end{thebibliography}

\end{document}